\documentclass[conference]{IEEEtran}
\IEEEoverridecommandlockouts

\ifCLASSINFOpdf
\else
\fi
\usepackage{epsfig}
\usepackage{graphicx}
\usepackage{xspace}
\usepackage{amssymb}

\makeatletter
\DeclareRobustCommand\onedot{\futurelet\@let@token\@onedot}
\def\@onedot{\ifx\@let@token.\else.\null\fi\xspace}

\def\ie{\emph{i.e}\onedot}

\def\etal{\emph{et al}\onedot}
\makeatother
\begin{document}

%
\title{Query by String word spotting based on character bi-gram indexing}

\author{\IEEEauthorblockN{Suman K. Ghosh and Ernest Valveny }
\thanks{$^{*}$This work is partially
supported by spanish project TIN2013-41751-P and a research grant from UAB}
\IEEEauthorblockN{Computer Vision Center, Dept. Ci\`{e}ncies de la Computaci\'o\\Universitat Aut\`{o}noma de Barcelona, 08193 Bellaterra (Barcelona), Spain \\
Email: sghosh,ernest@cvc.uab.es}}

\maketitle

\begin{abstract}
In this paper we propose a segmentation-free query by string word spotting method. Both the documents and query strings are encoded using a recently proposed word representation that projects images and strings into a common atribute space based on a pyramidal histogram of characters(PHOC). These attribute models are learned using linear SVMs over the Fisher Vector representation of the images along with the PHOC labels of the corresponding strings. In order to search through the whole page, document regions are indexed per character bi-gram using a similar attribute representation. On top of that, we propose an integral image representation of the document using a simplified version of the attribute model for efficient computation. Finally we introduce a re-ranking step in order to boost retrieval performance. We show state-of-the-art results for segmentation-free query by string word spotting in single-writer and multi-writer standard datasets.
\end{abstract}


%
\IEEEpeerreviewmaketitle

\section{Introduction}
 Enabling indexing and browsing over large handwritten databases is an elusive goal in document analysis. State of the art OCR technologies available for printed documents are not directly applicable to these type of documents due to challenges like diversity of the handwriting style or the presence of noise and distortion in historical manuscripts. This problem becomes even more challenging in multi-writer datasets.
 
 Word spotting has been proposed as an alternative to OCR, as a form of content-based retrieval procedure, which results in a ranked list of word images that are similar to the query word. The query can be either an example image (Query-By-Example (QBE)) or an string containing the word to be searched (Query-By-String (QBS)). Methods following QBE paradigm presents a huge disadvantage in practical applications as in order to spot a word the user needs to first locate/input an instance of such word. On the other hand QBS methods allow the user to type the keyword to search in a 
much more natural way.
 
 Initial approaches on word spotting followed a similar pipeline as OCR technologies, starting with binarization followed by structural/layout analysis and segmentation at word and/or character level. Example of this type framework are the works of \cite{Vinci,Serrano}. The main drawbacks of these methods come from the dependence on the segmentation step, which can be very sensible to handwriting distortions. Other initial attempts on QBS based methods relied on the extraction of letter or glyph templates, either manually \cite{Konidaris,Leydier} or by means of some clustering scheme \cite{Marinai,Liang}. Then these character templates are put together in order to synthetically generate an example of the sought
word. Although such methods proved to be effective and user-
friendly, their applicability is limited to scenarios where individual characters can be easily segmented. 
More generic solutions have been proposed in \cite{Fischer,Frinken}, where they learned models for individual characters and the relationship among them using either an an HMM \cite{Fischer} or a NN \cite{Frinken}. These models are trained on the whole word or even on complete text lines without needing an explicit character segmentation. They are used to generate a word model from the query string that has to be compared with the whole database at query time. Therefore, computational time can quickly increase with the size of the dataset. In this context it can be mentioned that example based methods are in a clear advantage as they can represent handwritten word images(queries) holistically by compact numeric feature vectors. 

One difficulty using a compact representation in QBS methods is that word strings and word images are not directly comparable. 
In a recent work \cite{almazan2014}  Almaz\'an \etal proposed a PHOC based attribute representation which can be used to represent both word images and strings. The attribute representation encodes the spatial position of characters in the word image through a Pyramidal Histogram of Characters (PHOC) and is learned using the powerful Fisher Vector representation of the images. Once word images are represented in this attribute space comparing images and strings is reduced to a Nearest Neighbour problem. Though this framework has achieved high accuracy in case of segmented words it can not be applied directly in a segmentation-free approach using sliding window protocol over entire document as it involves computation of costly fisher vector representation, unfeasible at query time.

In this work we propose to extend this approach to a segmentation-free scenario. To overcome the computational cost we propose to index different image regions based on the presence of character bi-grams (henceforth referred as only bi-grams in this article) to reduce the search space before using a sliding window protocol. Document images are segmented in different regions using a very basic segmentation method and then, for each bi-gram a ranked list of these regions is created. N-gram based language models have been widely used to improve OCR accuracy, but also for out-of-vocabulary word spotting in  \cite{Jawahar1} and for recognising complex degraded documents in \cite{Jawahar2}. However, our approach to using character bi-grams is more similar to inverted index files, a common practice in most successful text retrieval systems. At query time, given a query string, the indexing structure is accessed to retrieve all image regions containing any of the bi-grams in the string. Then, these regions are searched using a sliding window framework over the PHOC attribute representation. To achieve more computational efficiency, we  propose to pre-compute an integral image of the attribute representation. For that, some simplifications have to be done which makes the final attribute representation a bit less discriminative. To overcome this we propose an additional re-ranking step of the top candidate windows which uses the same attribute representation as of\cite{almazan2014}. 

Our main contributions can be summarized as: i) we propose an indexing scheme
 of document images based on character bi-grams. ii) We propose an efficient computation of the attribute word representation over a whole document using an integral image. iii) We combine an initial ranking based on the bi-gram indexing and integral image based word representation with a re-ranking step on the top candidate windows using a more powerful attribute representation. iv) We show results in a full segmentation-free scenario where, up to our knowledge, no previous results have been reported.

The rest of the paper is organized as follows: in section \ref{sec:MethodDesc} the proposed methodology is discussed in detail including index generation using bi-grams, computation of word attributes, retrieval and re-ranking. In section \ref{sec:Results}  experimental validation of the proposed method is discussed. Finally, we devote a section to conclude the paper with future directions of research.  

\begin{figure*}[!t]
\includegraphics[width=0.9\linewidth]{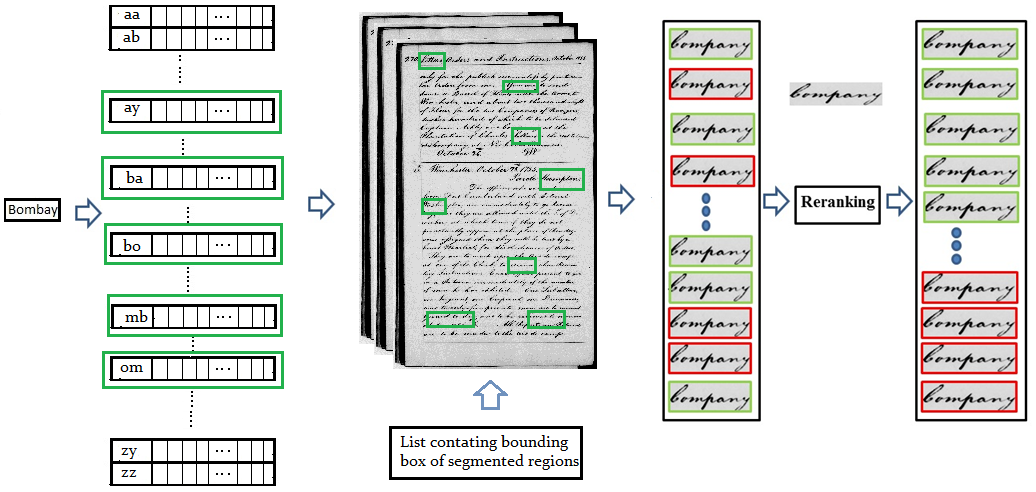}
\caption{General overview of the proposed pipeline}
\label{figure:schematicQBS}
\end{figure*}

\section{Method Description}
\label{sec:MethodDesc}
The proposed approach is illustrated in figure \ref{figure:schematicQBS}. The main idea is to index the regions over the document according to the presence of particular bi-grams in that region. At query time bi-grams present in the query word are identified and then the regions corresponding to the bi-grams are searched for presence of the query word. To identify the regions to be indexed according to bi-grams, a very naive segmentation based on connected component analysis is performed. To generate the index and to match candidate word images with text strings an attribute based representations similar to that of Almaz\'an\etal \cite{almazan2014} is used. In the following subsections we describe in detail each of the components of our approach: the attribute representation proposed in \cite{almazan2014}, efficient computation using an integral image, the indexing scheme, retrieval and re-ranking.



\subsection{Attribute Computation}
\label{sec:attributes-original}
In Almaz\'an\etal \cite{almazan2014} a common low dimensional representation is learned  for word images and text strings, that permits to address retrieval as a simple nearest neighbor problem. Though this representation can be utilized to accomplish both QBE and QBS, here our focus is on QBS word spotting. 

First, text strings are represented by a $d-$ dimensional binary embedding. This embedding -- called Pyramidal Histogram Of Characters (PHOC) -- encodes if a particular character appears in a particular spatial region of the string using a pyramidal decomposition that makes it more discriminative. The first level is just a basic histogram of characters encoding the presence or absence of a particular character in the string. Then, new levels  are added where at each level of the pyramid the word is further split and a new histogram of characters is added for each new division to account for characters at different parts of the word. In particullar Almaz\'an\etal \cite{almazan2014} used histograms at levels 2, 3, 4 and 5. In addition they also used a histogram of the 50 most popular bi-grams at level 2 thus resulting in a $604$ dimensional word representations.

Then, this embedding is used as a source for learning character attributes from word images. Each word image is projected into a $d-$dimensional space (same dimension as the PHOC representation) where each dimension is a character attribute. In a way similar to the PHOC string representation each characte attribute encodes the probability of appearance of a given character in a particular division of the image, using the same pyramidal decomposition as in the PHOC representation. Each attribute is independently learned using an SVM classifier on a Fisher Vector description of the word image, enriched with the $x$ and $y$ coordinates and the scale of the SIFT descriptor. 

More formally, given a training image $I$ and its associated text transcription, we can compute its Fisher Vector representation \cite{Peronnin} $f(I)$, where $f(I)$ is a function of the form  $f:I \to R^D$ being $D$ the dimension of the Fisher Vector representation. Now to project Fisher Vector representations into the PHOC attribute space, we learn an embedding function $\phi_I$ of the form $\phi_I:I \to R^d$ such that  \\
\begin{equation}
\hspace{15mm}\phi_I(I) = \mathbb{W}^{\mathbb{T}}f({I})
\end{equation}

\noindent{where $W$ is a matrix with an SVM-based classifier for each attribute, that are learned using the PHOC labels obtained from the text transcription of all the training words.}

At query time, text queries are encoded using the PHOC representation and word images are described with this attribute representation. Retrieval simply translates into finding the word candidates whose attribute representation is close to that of the query image. Almazan \etal \cite{almazan2014} proposed an additional step consisting of learning a common subspace between strings and images as direct comparison between PHOCs and attribute representations is not well defined since PHOCs are binary, while the attribute scores are not. Thus, a final calibration step is added, using Canonical Correlation Analysis, that aims at  maximizing the correlation among both representations. 

This final calibration and dimensionality reduction step can be represented with an additional embedding function $\psi$ represented as :$\psi_I:I \to R^{d'}$ and can be given as:
\begin{equation}
\hspace{15mm}\psi_I(I) = U^T\phi_I(I)
\end{equation}

\noindent{being $U$ the transformation matrix obtained with Canonical Correlation Analysis}

In this work, we have used this representation in the final retrieval step. For indexing, and relying on the same framework, we have defined an alternative representation based on bi-grams. Our starting hypothesis is that bi-grams can be discriminative word features that can be used as the basis for localizing areas of the document where the word is likely to appear. Thus, the goal of this new representation is to identify the presence of a particular bi-gram in a word image. To select the bi-grams that are used to generate the index over the handwritten documents, we refer to the study done by Jones \etal in \cite{Jones} where they studied a large corpus of 183
million words.
We consider 150 most popular bigrams from this study which covers 99.21\% of the total corpus.
To include numeric fields the digits from 0-9 are also included in this representation. Then, this particular representation is obtained using 150 bi-grams at level 2 of decomposition and using 10 digits at levels 2 and 3, thus resulting in a 350 dimensional representation. This representation is henceforth referred as PHOB (Pyramidal Histogram of Bi-grams) in this article. Using the strategy described above, a similar embedding function is learned to project the Fisher Vector representation of word images into the PHOB attribute space.

\subsection{Efficient computation of attributes using Integral Image}
\label{sec:integral-image}

In a scenario where the retrieval procedure has to rely on deciding among many (probably overlapping) candidate windows, the main bottleneck of using word attributes as basic representation is that it involves the redundant and costly computation of SIFT descriptors and Fisher Vector representation at run time -- it takes around $110$ms for a single candidate window --. To alleviate these problems we propose to pre-compute off-line the attribute representation for every pixel of the image and store it in an efficient integral image \cite{Dalal} that can be used to compute very fast the representation of any candidate window at query time. 

To describe the computation of the integral image of the attribute representation, let us denote the document images of the dataset as $I^k,k=1...n$ where n is the total number of images. For a given image $I^k$, we first compute the set of dense SIFT descriptors $d_{i,j}^k$ at every location $(i,j)$. Then, we can define the embedding function into the attribute space $\phi_I$ for every pixel location as:
\begin{equation}
\hspace{15mm}\phi_I(i,j) = \mathbb{W}^{\mathbb{T}}f({d_{i,j}})
\end{equation}
\noindent{where $f({d_{i,j}})$ is the fisher vector representation for the $(i,j)$ pixel of image $I^k$ and $W$ is a matrix encoding the attribute classifiers as in previous section. Finally this attribute representation for every pixel is projected to the lower dimensional subspace obtained through Canonical Correlation Analyisis using the same transformation matrix $U$ introduced in previous section:}
\begin{equation}
\hspace{15mm}\psi_I(i,j) = \it{\mathbb{U}^{\mathbb{T}}\phi_I(i,j)}
\end{equation}



Once we have the final attribute representation for every pixel, it can be easily aggregated into an integral image $\Psi_{i,j}$: 
\begin{equation}
\hspace{15mm}\Psi_{i,j} = \sum_{i^{'}<=i, j^{'}<=j}{\psi_{i,j}}
\end{equation}

The time and memory requirements for computing the attribute representation representation can be further reduced if we arrange the image into $NXN$ dimensional blocks and instead of computing Fisher Vector representation for every pixel, we only compute one Fisher Vector for each block. 

Although we end up obtaining an integral image encoding an attribute-based representation very similar to that of \cite{almazan2014}, in the process we have to apply some simplifications: i) as we are computing Fisher Vector on a per pixel basis, we can not have, at the time of computing the integral image, the relative position of the key-points inside a given candidate box. Therefore, SIFT descriptors cannot be enriched using the relative positional information $x$, $y$ coordinates, as explained in section \ref{sec:attributes-original}. ii) Also, as we cannot know the size of the underlying window can not apply the window size normalization performed in the original approach. These simplifications make the final representation a bit less discriminative. In section \ref{sec:Reranking} we will introduce a re-ranking step to compensate this loss of disciminability.

\subsection{Index generation} 
\label{sec:Indexing}

The goal of the indexing scheme is to create an ordered list of regions per bi-gram over the entire document database, so that at query time only regions relevant to bi-grams in the query word have to be searched. This index file is similar to inverted index files used frequently in text retrieval. 

To generate a list of regions for each bi-gram, the document is first segmented. Please note that the goal of segmentation is only to identify regions of plausible occurrence of bi-grams in documents not to find an exact and accurate word segmentation.

\subsubsection{Segmentation}
For segmentation, the document image is first binarized by setting the threshold as 75\% of the mean intensity of the image. Then, each document in the database is represented as a set of connected components. 
Connected components which are overlapping with each other are merged into a single region. However in english handwriting some descendant of a line can sometimes overlap with ascendants of the line below. Thus merging connected component sometimes can merge components from two different lines.
To avoid this a minimal bounding box for each connected component is found(by greedily finding the smallest region that
contains
90\%
of the density of the binarized image).
To find overlapping connected components, instead of actual bounding box this minimal bounding box is used. Connected components which are very close along width of the document page are also merged together to form a single region. However this notion of closeness can vary from one document to another. To overcome this we use different thresholds thus yielding one set of regions per threshold after merging. At the end we merge all of these regions to obtain the list of regions which are used for indexing.

\subsubsection{Indexing by bi-grams} 
To generate index over the document database represented by the segmented regions, bi-grams are considered as strings and represented by the PHOB representation introduced in section \ref{sec:attributes-original}. Each segmented region is also represented by the corresponding PHOB based attribute representation using the  Fisher vector of the region.
Both representations are then converted to a low dimensional common subspace using CCA calibration. Similarity between bi-grams and segmented regions can be computed as a dot product between their corresponding representations in this space. For each bi-gram, segmented regions are sorted in order of decreasing similarity and stored as inverted index files.

\subsection{Retrieval}
\label{sec:retrieval}
At retrieval time, given a text string the aim is to extract all the occurrences of the string in the entire dataset.
For that, first all distinct bi-grams of the text string are found. Then, for each bi-gram the inverted index is searched and top $n$ regions for each one are further considered for retrieval. Thus, for a query having $k$ distinct bi-grams, $k\times n$ regions are searched. However many regions can be indexed by more then one bi-gram thus making the number of distinct regions to search much less for most cases.

Once potential regions are identified, we employ a sliding window search using the integral image representation described in section \ref{sec:integral-image}. We apply the sliding window over the region as returned by the index file if the region is big enough to accommodate the query word as calculated by the mean width of all the same training words. If the region is small then we merge it with the regions on the left and on the right to make a bigger region where the sliding window can be employed. Using the integral image, the attribute-based representation of each candidate window explored by the sliding window can be easily obtained and compared through the cosine similarity with the PHOC representation of the query string. Finally, the candidate window list is ranked in order of decreasing similarity and non maximal suppression is performed to obtain the final relevance list.

\subsection{Re-ranking}
\label{sec:Reranking}
As we have explained in section \ref{sec:integral-image} computing offline the integral image of attributes before query time requires certain simplifications with respect to the original attribute representation. Though these simplifications make the overall procedure efficient and faster to execute, they also make the representation less discriminative leading to a significant loss in accuracy. In order to alleviate this effect, and similar to other applications in image retrieval \cite{Chum,Arand} a re-ranking step is introduced. Basically it consists of applying more discriminative and costly features to the best retrieved windows by the first ranking step in order to obtain the final ranking list. In word spotting, \cite{almazan2014a} applied re-ranking by using  Fisher Vector as a second ranking step after selecting windows using a HOG based representation.

In this work we use  the same strategy: the top  $N\%$ candidates from the ranked list given by the initial ranking obtained with the sliding window search are re-ranked using the more discriminative original attribute representation described in section \ref{sec:attributes-original}. 
\begin{table*}
\caption{Result of our word spotting method in comparison with  state-of-the-art. (1) Proposed method without the reranking step. (2) Re-ranking with top 60\% of candidates. (3) Re-ranking with top 80\% of the candidates from the first step. }
\label{table:resMap}
\begin{center}
\begin{tabular}{l|cccc}
\hline
 & segmentation&Queries&GW &IAM \\
\hline\rule{0pt}{12pt}

  Proposed (1)  &  -&All queries& 64.32&39.47 \\
Proposed (2) with RR (60\%)  &  -& All queries& 69.87&43.78\\ Proposed (3) with RR (80\%)  &-&All queries&   73.7&48.57\\
\hline\rule{0pt}{12pt}
Liang et al.\cite{Liang} &Word-level& 38 queries&67\%
at rank 10\\
Fischer et al.\cite{Fischer} &Line-level&In-vocabulary words
&62.
08&47.75\\
Frinken et al. \cite{Frinken} &Line-level &In-vocabulary words&
71&76\\
\hline
\end{tabular}
\end{center}
\end{table*}
\section{Experimental Results}
\label{sec:Results}
The proposed method have been evaluated two publicly available databases widely used in state-of-the art word spotting systems.

One of them is a single writer handwritten database popularly known as \textbf{The George Washington (GW) dataset},comprising 5000 words annotated at word level from 20 handwritten letters written by George Washington to his associates in 18th century.

To evaluate our method in a multi-writer scenario \textbf{The  IAM Offline Dataset}\cite{Marti1} is used. It is a large dataset
comprised of 1539 pages of modern handwritten English text written by 657
different writers. The document images are annotated at word and line level
and contain the transcriptions of more than 13000 lines and 115000 words. We follow the official partition for writer independent text line recognition task.  



To evaluate the proposed method every unique ground truth text in the GW dataset is considered as a query and after ranking a candidate window is considered as a true positive if it overlaps by more then 50\% with any of the ground truth annotated boxes of the same word. We measure accuracy in terms of Mean Average Precision. This protocol is in line with most of the segmentation-free word spotting works such as \cite{almazan2014}. 

In the case of the IAM dataset however, we follow a different strategy, performing line spotting instead of word spotting, \ie the whole lines are retrieved if they contain the query word. Each query word is searched inside all annotated text lines. The distance between the query and a given text line is defined as the distance between the query and the closest candidate word of that line. A similar strategy has been followed by Almaz\'an \etal in  \cite{almazan2014}. 

Table \ref{table:resMap} summarizes the results of our method. We provide results for three settings of our method: the first one without re-ranking, and then re-ranking with two different choices of $N$. Table \ref{table:resMap} also reports the performance of other similar works reported in \cite{Liang,Fischer,Frinken}. However direct comparison with these methods is not straightforward, as they use different protocols to evaluate their methods: both \cite{Fischer,Frinken} compute the average precision at line level and only use in-vocabulary words in the evaluation. Besides they do not perform full segmentation-free word spotting as they require a precise line segmentation.  In \cite{Liang} Liang et al. used a selected set of queries to obtain the mAP just considering the first 10 retrieved elements. From the results it can be observed that results of our baseline system without any re-ranking is slightly poorer than Liang \etal \cite{Liang}. With re-ranking we obtain the best results among all in the GW dataset.  In the IAM dataset our system also gives better result than that of Fisher \etal \cite{Fischer}, but performs poorer than that of Frinken \etal \cite{Frinken}. However, fair comparison with this method is difficult due to differences in the method and the evaluation protocol.



Average computational time to evaluate a query is an important criteria for any word spotting system in real life applications. Unfortunately, none of the methods used for comparison reported computational time. Thus comparison is not possible but it is worth mentioning that it takes about 1.45s per query o
n average to evaluate a query using our method without re-ranking for the GW dataset. However this computational time increases to 8.63s per query when top 80\% candidates are re-ranked. It can be observed that the proposed method without the re-ranking step is quite fast to be used in a real time environment. The re-ranking step significantly increases the computational time as it must compute SIFT and Fisher Vector for each candidate window.

\section{Conclusion}

This paper proposes a segmentation-free approach to word spotting using QBS in document images. To reduce computational cost of comparing a large number of candidates an inverted index per character bi-grams is generated over the entire document offline. At query time a query word is searched only in these indexed regions. Both query strings and candidate words are represented using PHOC based compact numeric feature vector.
An efficient way to compute PHOC based word attributes in query time by using pre-computed integral image representation is also shown. The results of our method shows significant improvements over the current state-of-the-art. Computational time could be further improved integrating our approach with a compression technique such as product quantization as done in \cite{almazan2014a}. The proposed method is based on a simplification of the original attribute word representation. Context information around a pixel can be exploited in the future in order to compensate for the poorer Fisher Vector representation that we use in our method. 


\begin{thebibliography}{1}

  


\bibitem {Vinci}

 A. Vinciarelli, S. Bengio, H. Bunke, Offline recognition of unconstrained handwritten texts using HMMs and statistical language models, IEEE Transactions on PAMI, 26 (2004) 709–-720.
 
 \bibitem {Serrano}
J. Rodr\'{\i}guez-Serrano, F. Perronnin, Local gradient histogram features for word spotting in unconstrained hand-written documents: ICFHR, 2008







\bibitem{almazan2014a}
J. Almaz\'an and A. Gordo and A. Forn\'es and E. Valveny,
Segmentation-free Word Spotting with Exemplar SVMs,
Pattern Recognition,2014

\bibitem{almazan2014} 
J. Almaz\'an and A. Gordo and A. Forn\'es and E. Valveny, 
Word Spotting and Recognition with Embedded Attributes, 
TPAMI,2014


\bibitem{Marti1}
 U.-V. Marti and H. Bunke, The IAM-database: An english sentence database for off-line handwriting recognition,
IJDAR,
2002.



\bibitem{Chum}
O. Chum, J. Philbin, J. Sivic, M. Isard, A. Zisserman, Total recall: Automatic query expansion with a generative feature model for object retrieval, in ICCV, 2007, pp. 1–-8.

\bibitem{Arand}
R. Arandjelovi \'c, A. Zisserman, Three things everyone should know to improve object retrieval, in IEEE CVPR, 2012, pp. 2911-–2918.

\bibitem{Peronnin}
 F. Perronnin, J. S\'anchez, and T. Mensink,Improving the Fisher kernel for large-scale image classification,in ECCV,2010.

\bibitem{Dalal}
N. Dalal and B. Triggs. Histograms of oriented gradients for human detection.in IEEE CVPR, 2005.

\bibitem{Kovalchuk}
Alon Kovalchuk, Lior Wolf, and Nachum Dershowitz
A Simple and Fast Word Spotting Method,in ICFHR 2014



\bibitem{Marinai} 
S. Marinai, E. Marino, and G. Soda, Font adaptive word indexing of modern printed documents, IEEE Transactions on Pattern Analysis and Machine Intelligence, vol. 28, no.8, pp. 1187–1199, August 2006.

\bibitem{Konidaris}
T. Konidaris, B. Gatos, K. Ntzios, I. Pratikakis, S. Theodoridis, and
S. Perantonis, Keyword-guided word spotting in historical printed doc-
uments using synthetic data and user feedback,International Journal
on Document Analysis and Recognition, vol. 9, no. 2–4, pp. 167–177,April 2007.

\bibitem{Leydier}
Y. Leydier, A. Ouji, F. LeBourgeois, and H. Emptoz, Towards an
omnilingual word retrieval system for ancient manuscripts,Pattern Recognition, vol. 42, no. 9, pp. 2089–2105, September 2009.

\bibitem{Liang}
Y. Liang, M. Fairhurst, and R. Guest, A synthesised word approach to word retrieval in handwritten documents, Pattern Recognition, vol. 45, no. 12, pp. 4225–4236, December 2012.

\bibitem{Fischer} 
A. Fischer, A. Keller, V. Frinken, and H. Bunke, Lexicon-free hand-
written word spotting using character HMMs,Pattern Recognition
Letters
, vol. 33, no. 7, pp. 934–942, May 2012.

\bibitem{Frinken} V. Frinken, A. Fischer, R. Manmatha, and H. Bunke, A novel word
spotting method based on recurrent neural networks,
IEEE Transac-
tions on Pattern Analysis and Machine Intelligence
, vol. 34, no. 2, pp.
211–224, February 2012
\bibitem{Jawahar1}
Udit Roy, Naveen Sankaran, K. Pramod Sankar and C. V. Jawahar,Character N-Gram Spotting on Handwritten Documents Using Weakly-Supervised Segmentation,Proceedings of the 2013 12th International Conference on Document Analysis and Recognition
Pages 577-581 

\bibitem{Jawahar2}
Shrey Dutta, Naveen Sankaran, K Pramod Sankar and  CV Jawahar,Robust recognition of degraded documents using character n-grams, Proccesdings of 10th IAPR International Workshop on Document Analysis Systems (DAS),pages 130-134

\bibitem{Jones}
Michael N. Jones and D. J. K. Mewor, Case-sensitive letter and bigram frequency counts
from large-scale English corpora, Behavior Research Methods, Instruments \& Computers
August 2004, Volume 36, Issue 3, pp 388-396 

\end{thebibliography}
\end{document}